\pdfoutput=1
\documentclass[lettersize,journal]{IEEEtran}
\usepackage[T1]{fontenc}    % use 8-bit T1 fonts
\usepackage{hyperref}       % hyperlinks
\usepackage{url}            % simple URL typesetting
\usepackage{booktabs}       % professional-quality tables
\usepackage{amsfonts}       % blackboard math symbols
\usepackage{nicefrac}       % compact symbols for 1/2, etc.
\usepackage{microtype}      % microtypography
\usepackage{xcolor}         % colors
\usepackage{gensymb}
\usepackage{multirow}
\usepackage{makecell}
\usepackage{array}
\usepackage{graphicx}
\usepackage{amsmath}
\usepackage{float}
\usepackage{subfig}

\usepackage[table]{xcolor}

\usepackage{textcomp}
\usepackage{xspace}
\newcommand{\method}{VideoTG-R1\xspace}

\newcommand{\eg}{\textit{e.g.}}
\newcommand{\ie}{\textit{i.e.}}

\ifCLASSINFOpdf
\else
\fi
\begin{document}
	\title{VideoTG-R1: Boosting Video Temporal Grounding via  Curriculum Reinforcement Learning on Reflected Boundary Annotations}

	\author{Lu Dong, Haiyu Zhang, Han Lin, Ziang Yan, Xiangyu Zeng, Hongjie Zhang, Yifei Huang, Yi Wang,~\IEEEmembership{Member,~IEEE},  Zhen-Hua Ling,~\IEEEmembership{Senior Member,~IEEE},  Limin Wang,~\IEEEmembership{Member,~IEEE}, Yali Wang
		% <-this % stops a space
		% \thanks{This paper was supported in part by XXX.}% <-this % stops a space
		\thanks{Lu Dong is with University of Science and Technology of China, Hefei 230027, China, and also with Shanghai Artificial Intelligence Laboratory, Shanghai 202150, China. (Email: dl1111@mail.ustc.edu.cn)}
            \thanks{Haiyu Zhang is with  Beihang University, Beijing 100191, China, and also with Shanghai Artificial Intelligence Laboratory, Shanghai 202150, China. (Email: zhyzhy@buaa.edu.cn)}
            \thanks{Han Lin is with Shanghai Jiao Tong University, Shanghai 200240, China, and also with Shanghai Artificial Intelligence Laboratory, Shanghai 202150, China. (Email:  linhan1201@sjtu.edu.cn)}
            \thanks{Ziang Yan is with Zhejiang University, Hangzhou 310058, China, and also with Shanghai Artificial Intelligence Laboratory, Shanghai 202150, China. (Email: yanziang@pjlab.org.cn)}
           \thanks{Xiangyu Zeng and Limin Wang are with State Key Laboratory for Novel Software Technology, Nanjing University, Nanjing 210023, China, and also with Shanghai Artificial Intelligence Laboratory, Shanghai 202150, China. (Email: XiangyuZeng2001@outlook.com and lmwang@nju.edu.cn)}
            %  XiangyuZeng2001@outlook.com 
            \thanks{Hongjie Zhang, Yifei Huang,   and Yi Wang are with Shanghai Artificial Intelligence Laboratory, Shanghai 202150, China. (Email: nju.zhanghongjie@gmail.com, hyf015@gmail.com,  and wangyi@pjlab.org.cn)}
            \thanks{Zhen-Hua Ling is with University of Science and Technology of China, Hefei 230027, China (Email: zhling@ustc.edu.cn)}
            
            \thanks{Yali Wang is with Shenzhen Institute of Advanced Technology, Chinese Academy of Sciences, Shenzhen 518055, China, and also with Shanghai Artificial Intelligence Laboratory, Shanghai 202150, China. (Email: yl.wang@siat.ac.cn).
            \textit{Yali Wang is the corresponding author}.}
	}
	
	% \author{Xinyu Xu, Huazhen Liu, Feiming Wei, Huilin Xiong, \\ Wenxian Yu,~\IEEEmembership{Senior Member,~IEEE}, and Tao Zhang$^{\ast}$ \thanks{*Corresponding author},~\IEEEmembership{Member,~IEEE},
	% 	% <-this % stops a space
	% 	\thanks{This paper was supported in part by XXX.}% <-this % stops a space
	% 	\thanks{X. Xu, T. Zhang, F. Wei, H. Xiong, and W. Yu are with the Shanghai Key Laboratory of Intelligent Sensing and Recognition, School of Sensing Science and Engineering, Shanghai Jiao Tong University, Shanghai 200240, China.}
	% 	\thanks{H. Liu is with the Intelligent Photoelectric Sensing Institute, School of Sensing Science and Engineering, Shanghai Jiao Tong University, Shanghai 200240, China.}
	% }

	\markboth{Submission to IEEE Transactions on Circuits and Systems for Video Technology}%
	{Shell \MakeLowercase{\textit{et al.}}: Bare Demo of IEEEtran.cls for IEEE Transactions on Magnetics Journals}



	% for Transactions on Magnetics papers, we must declare the abstract and
	% index terms PRIOR to the title within the \IEEEtitleabstractindextext
	% IEEEtran command as these need to go into the title area created by
	\IEEEtitleabstractindextext{%
		\begin{abstract}
			Video temporal grounding (VTG) aims to locate precise segments in videos based on language queries, which is a fundamental challenge in video understanding. While recent Multimodal Large Language Models (MLLMs) have shown promise in tackling VTG through reinforcement learning (RL), they overlook
the challenges arising from both the quality and difficulty of training
samples. (1) \textbf{Partially annotated samples}.  Many samples contain relevant segments beyond the annotated interval, introducing ambiguous supervision. (2) \textbf{Hard-to-ground samples}. Samples with poor zero-shot performance produce consistently low and indistinguishable rewards during RL training, exhibiting no clear preference among multiple outputs and thus hindering learning efficiency. To address these challenges, we propose \method, a novel curriculum RL framework with reflected boundary annotations, enabling data-efficient training. Specifically, we propose a Boundary Reflection Agent that utilizes MLLMs to predict query-relevant timestamps outside the annotated intervals, allowing us to identify and filter out partially annotated samples, thereby reducing ambiguity. Furthermore, we introduce a Difficulty Estimation Agent to assess the training difficulty of each sample and design a curriculum RL strategy that
dynamically masks the videos of hard-to-ground samples according to the training steps, easing the training difficulty and providing clearer preference.
Experiments on the VTG and grounded VideoQA tasks demonstrate the effectiveness of our method. 
Remarkably, with only 10\% of the training samples and 21\% of the computational budget, VideoTG-R1 outperforms full-data counterparts under both group relative policy optimization (GRPO) and supervised fine-tuning (SFT). The code is available at https://github.com/ldong1111/VideoTG-R1.

    \end{abstract}
		
		}

	\maketitle

	\IEEEdisplaynontitleabstractindextext

	% \fi
	\IEEEpeerreviewmaketitle

	\section{Introduction}

    % \begin{figure*}[t] % [h] 代表尽量放在当前位置
%     \centering
%     % linewidth
%     \includegraphics[width=1.0\textwidth]{AnonymousSubmission/LaTeX/figs/0718_framework_V3.pdf} % 调整图片宽度
%     \vspace{-4mm}
%     \label{fig:motivation} % 为图片设置引用标签
% \end{figure*}
\begin{figure*}[t] % [h] 代表尽量放在当前位置
    \centering
    % linewidth
    \includegraphics[width=1.0\linewidth]{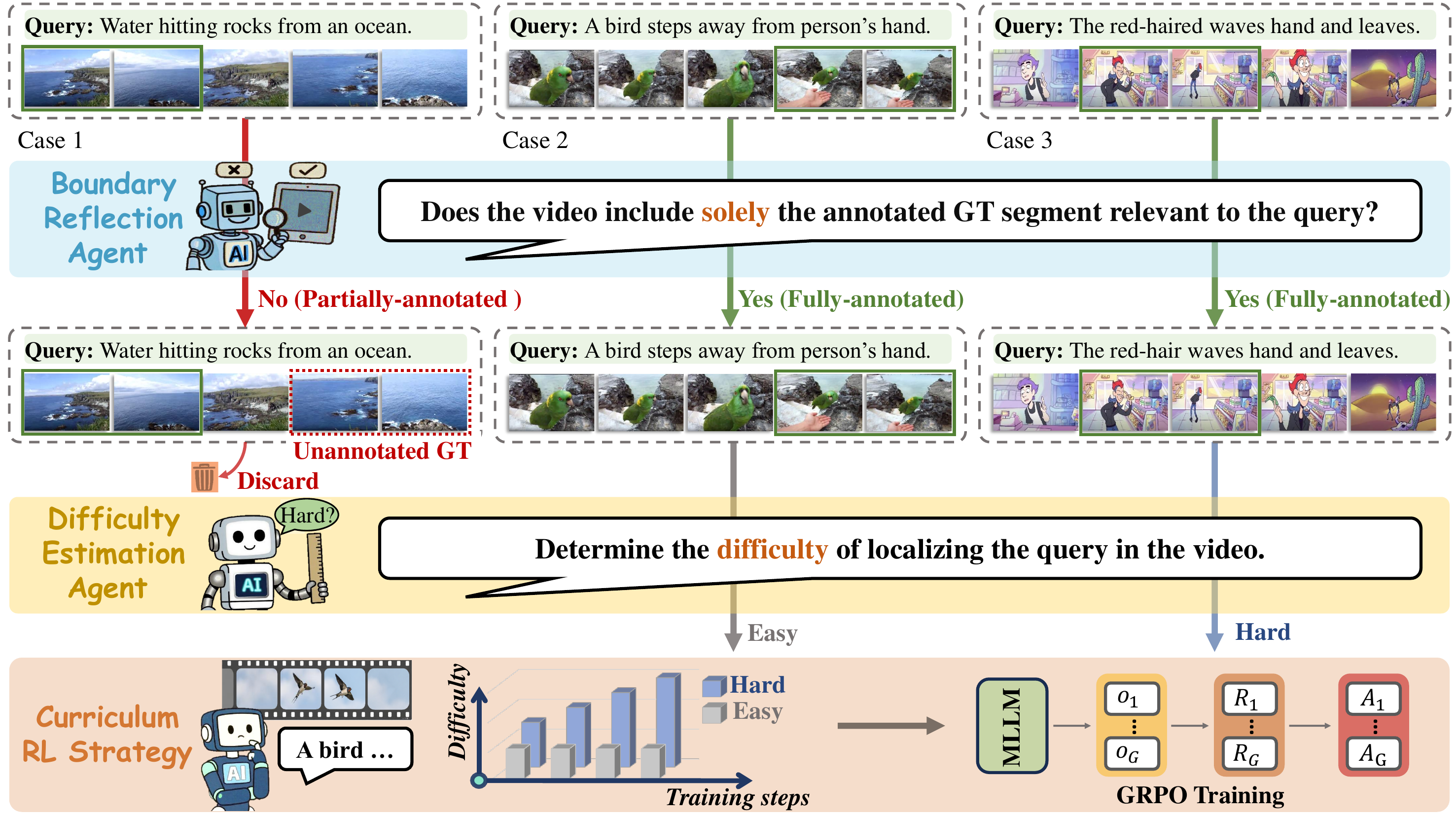} % 调整图片宽度
    \vspace{-4mm}
    \caption{\textbf{Overview.} \method is a multi-agent system for efficient RL training. It contains three key modules to address the primary challenges in VTG. Boundary Reflection Agent quantifies the missing annotations and identifies the partially annotated samples. Difficulty Estimation Agent estimate the difficulty of each sample via zero-shot evaluation. Curriculum-RL strategy learns the hard samples in an easy-to-hard manner by dynamically masking the videos.
    }
    \label{fig:motivation} % 为图片设置引用标签
\end{figure*}

	\IEEEPARstart{U}{nderstanding} long-form videos is a core ambition in computer vision~\cite{gaidon2013temporal,comanici2025gemini}. A fundamental step toward this goal is Video Temporal Grounding (VTG)~\cite{gao2017tall},
which aims to pinpoint the event timestamp within a video based on the given query.
VTG has a wide range of applications, including smart home assistants~\cite{yang2025egolife}, video retrieval~\cite{caba2015activitynet}, and human-computer interaction systems~\cite{huang2020mutual}. 

Traditional VTG approaches typically adopt a two-stage feature-based pipeline~\cite{liu2022memory,liu2022reducing,sun2023video,zhang2023video,qi2024collaborative,dong2025weakly}, which first extracts features with pretrained models and then adopts task-specific grounding models to predict the timestamps. While effective, the task-specific design often fails to generalize in a zero-shot setting, which is crucial for real-world scenarios~\cite{huang2024vtimellm}. Recently, Multimodal Large Language Models (MLLMs)~\cite{huang2024vtimellm,huang2024lita,liu2025videomind} have emerged as versatile generalists, addressing VTG  tasks directly in a zero-shot manner by their unified, multitask reasoning capabilities. However, MLLMs tend to over-penalize false negatives during supervised fine-tuning (SFT)~\cite{liu2025visual}. Inspired by the strong reasoning capabilities exhibited by reinforcement-learning (RL)~\cite{shao2024deepseekmath,guo2025deepseek} in LLMs, researchers have begun to explore RL-based frameworks to effectively post-train MLLMs for VTG~\cite{li2025videochat}.

Despite the potential of RL for VTG, current approaches 
overlook the challenges arising from both the quality and difficulty of training samples.
(1) \textbf{Partially annotated samples}. Many samples contain relevant segments beyond annotated intervals. As shown in Case 1 in Fig.~\ref{fig:motivation}, the query ``\textit{Water hitting the rocks from an ocean}'' appears at both the beginning and the end of the video, but the end segment is unannotated, \ie, this sample's ground-truths are partially annotated. These partially annotated samples introduce ambiguous supervision, which misleads the optimization process and degrades the final performance. (2) \textbf{Hard‐to‐ground samples}. 
Samples with poor zero-shot grounding performance produce consistently low and indistinguishable rewards during RL training,
exhibiting no clear preference among multiple outputs and thus hindering the learning efficiency of RL.

To address these two issues, we introduce VideoTG-R1, 
a novel curriculum reinforcement learning framework with reflected boundary annotations to enable more data-efficient training.
Specifically, we propose a Boundary Reflection Agent that 
utilizes  MLLMs to predict query-relevant timestamps outside the annotated intervals, allowing us to identify and filter out partially annotated samples, thereby mitigating the ambiguous supervision. Additionally, we employ a Difficulty Estimation Agent to assess the training difficulty of each sample and design a curriculum RL strategy that dynamically masks the videos of hard-to-ground samples according to the training steps. This approach eases the training difficulty of hard-to-ground samples, provides clearer preference during the RL training progress, and facilitates better convergence.

In summary, our contributions are threefold: 

\begin{itemize}
\item We propose a novel \method framework to address two major challenges during the RL post-training stage in VTG, \ie, partially annotated samples and hard-to-ground samples.

\item We propose a Boundary Reflection 
 Agent to mitigate the ambiguous supervision caused by partially annotated samples, and introduce a Difficulty Estimation Agent along with a curriculum RL strategy to alleviate the training difficulty of hard-to-ground samples.
\item We validate \method through comprehensive evaluation on  VTG and grounded Video Question Answering (VideoQA). We achieve state-of-the-art performance. Moreover, with only 10\% of the training samples and 21\% of the computational budget, our method outperforms full-data counterparts under both GRPO and SFT paradigms. 

\end{itemize}

	\section{Related Work}

\subsection{Video Temporal Grounding}
The ability to decompose and recognize salient content segments within  videos~\cite{zhou2022thinking,wang2021exploring,hu2023learning,zhang2023cross,zhao2024snippets,shao2024text,yan2025task,zhao2025constructing,liu2025brtal} is a fundamental requirement for video understanding, reflecting a deep comprehension of both the temporal structure and semantic content. A key task within this domain is Video Temporal Grounding (VTG)~\cite{gao2017tall, krishna2017dense,sun2023video,zhang2023video,qi2024collaborative,dong2025weakly}, which aims to localize natural‐language queries within untrimmed videos. Early VTG approaches adopt feature-based pipelines. They first extract video and language features using pre-trained encoders, then predict timestamps based on multimodal interaction. These can be categorized into proposal-free~\cite{liu2022memory,liu2022reducing,lei2021detecting,jang2023knowing} and proposal-based~\cite{zhang2020learning,wang2022negative,lin2023univtg} methods. CRNet~\cite{sun2023video} proposes a multi-choice generator to generate favorable representations for each candidate moment.
DMFAT~\cite{zhang2023video} systematically analyzes and mitigates both visual and combinatorial biases in the training data without using a proposal-generation module. While effective, these approaches are difficult to generalize to zero-shot scenarios.
Recent works have resorted to MLLMs~\cite{huang2024vtimellm,huang2024lita, ren2024timechat,zeng2024timesuite,yan2025task,liu2024r,wu2025number,liu2025videomind}  for their unified, multitask reasoning capabilities.
VTimeLLM~\cite{huang2024vtimellm} 
proposes a boundary-aware three-stage training strategy to progressively integrate temporal information into MLLMs.
However, these methods tend to over-penalize false negatives during SFT~\cite{liu2025visual}. In this work, we focus on RL to improve the capability of MLLMs in the VTG task.

\subsection{RL in VTG}
Recent works~\cite{jaech2024openai, guo2025deepseek}  have significantly advanced LLM reasoning capabilities via reinforcement learning. For VTG, several methods~\cite{li2025videochat, park2025deepvideo}  adopt RL with rigorously designed rewards to enhance VTG performance. 
However, the impact of data quality and how to train hard-to-ground samples effectively have been underexplored. In this work, we enhance RL for VTG by filtering out partially annotated samples and employing a curriculum RL strategy to ease the training difficulty of hard-to-ground samples.

\subsection{Data Selection in MLLMs}
Low-quality samples in instruction datasets can lead to low training efficiency and degrade performance. Most previous works~\cite{gadre2023datacomp,lee2024concept,ma2025mllm,wang2025sota} aim to curate a subset of data that maximizes model performance while minimizing unnecessary computation.
For instance, COINCIDE~\cite{lee2024concept} utilizes inter-cluster transferability and density to represent the sample quality. 
These methods emphasize data diversity and representativeness, but often lack task-specific interpretability. In this work, we filter out partially annotated samples to avoid ambiguity in supervision, specifically tailored for VTG and offering strong interpretability. Regarding training efficiency, recent works~\cite{xiao2025fast,park2025deepvideo} have concentrated on tasks with discrete reward signals, \eg, ImageQA and VideoQA.~\cite{xiao2025fast} unlocks the model’s potential by tackling the hardest samples and then gradually transitioning to easier ones.  However, the training efficiency of VTG with contiguous reward signals has been less explored. In this work, we aim to ease the training difficulty of hard-to-ground samples for VTG.

	\section{Method}

In this section, we explain in detail the two key components of \method in Fig.~\ref{fig:motivation}. First, we present Boundary Reflection Agent to identify and filter out partially annotated samples (Sec.~\ref{sec:BRA}). Second, we propose Curriculum Reinforcement Learning (RL) with  a Difficulty Estimation Agent and a Curriculum RL strategy, where the Difficulty Estimation Agent assesses the training difficulty of each sample and a Curriculum RL strategy to provide clearer preference during RL training (Sec.~\ref{sec:CRL}).

\subsection{Boundary Reflection Agent}
\label{sec:BRA}

\begin{figure}[t] % [h] 代表尽量放在当前位置
    \centering
    \includegraphics[width=1.0\linewidth]{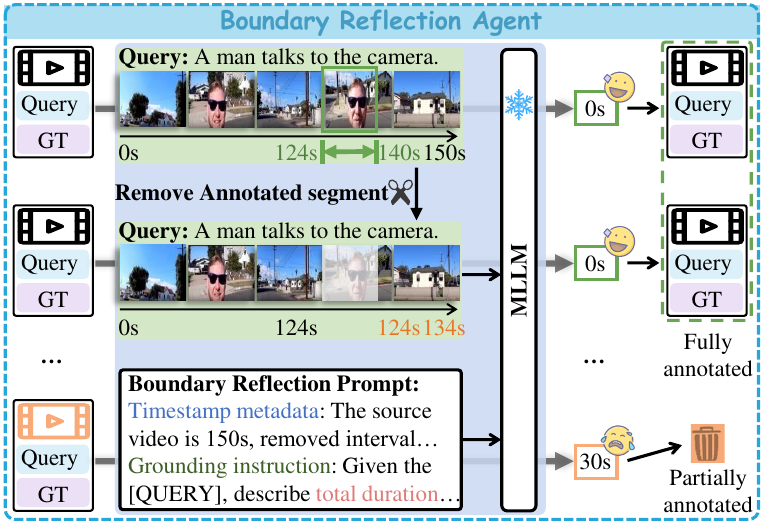} % 调整图片宽度
    \vspace{-4mm}
    \caption{\textbf{Boundary Reflection Agent}. First, the annotated segments are removed from the original video. Then, a boundary reflection prompt augmented with a timestamp metadata and a grounding instruction is fed to an MLLM to estimate the total duration of query-relevant segments outside the annotated video. Finally, we can identify and discard partially annotated samples.}
    
    % Finally, we discard any partially annotated samples in which the query occurs outside the annotated video.  }
    \label{fig:framework_bfa} % 为图片设置引用标签
\end{figure}

To address the issue of partially annotated samples in video temporal grounding datasets, we introduce a Boundary Reflection Agent. This agent identifies and filters out partially annotated samples to curate a cleaner subset, thereby facilitating more data-efficient training. The pipeline is shown in Fig.~\ref{fig:framework_bfa}. To be concrete, given a dataset $\mathcal{D}$ of $N$ samples,  each sample $\mathcal{S} = (\mathcal{Q}, \mathcal{V}, \mathcal{T})$ consists of a query $\mathcal{Q}$, a video 
$\mathcal{V}$ and ground truth (GT) timestamps $\mathcal{T}$. We employ a Boundary Reflection Agent to identify a subset $\mathcal{D}_{c} \subseteq \mathcal{D}$ that only contains fully annotated samples.

\subsubsection{Boundary Reflection}

% \begin{figure*}[t] % [h] 代表尽量放在当前位置
%     \centering
%     \includegraphics[width=1.0\textwidth]{figures/0728_full_framework_zhengrong.pdf} % 调整图片宽度
%     \vspace{-4mm}
%     \caption{Framework.  1) Boundary Reflection Agent. First, the GT interval is excluded from the raw video. Then, a boundary reflection prompt augmented with timestamp metadata and a grounding prompt is fed to MLLM to estimate the total duration of query-relevant segments outside the annotated video.   2) Difficulty Estimation Agent. We perform zero-shot evaluation with an MLLM to estimate the difficulty of each sample, and then split the dataset into ``hard'' and ``easy'' samples based on their predicted IoUs.  3) Curriculum RL strategy. We apply a dynamical mask that occludes segments outside the annotated ground-truth videos of hard samples to reduce the difficulty of hard samples.
%     }
%     \label{fig:framework} % 为图片设置引用标签
% \end{figure*}

% \begin{figure}[t] % [h] 代表尽量放在当前位置
%     \centering
%     \includegraphics[width=1.0\linewidth]{AnonymousSubmission/LaTeX/figs/0725_curriculum_augmentation.pdf} % 调整图片宽度
%     \vspace{-4mm}
%     \caption{a) Difficulty Estimation Agent. We perform zero-shot evaluation with an MLLM to estimate the difficulty of each sample, and then split the dataset into ``hard'' and ``easy'' samples based on their predicted IoUs.  b) Curriculum RL strategy. We apply a dynamical mask that occludes segments outside the annotated ground-truth videos of hard samples to reduce the difficulty of hard samples.
%     }
%     \label{fig:framework} % 为图片设置引用标签
% \end{figure}
% \begin{figure*}[t] % [h] 代表尽量放在当前位置
%     \centering
%     \includegraphics[width=1.0\textwidth]{AnonymousSubmission/LaTeX/figs/0801_curriculum_RL.pdf} % 调整图片宽度
%     \vspace{-4mm}
%     \caption{1) \textbf{Difficulty Estimation Agent}. We perform zero-shot evaluation with an MLLM to estimate the difficulty of each sample, and then split the dataset into ``hard'' and ``easy'' samples based on their predicted IoUs.  2) \textbf{Curriculum RL strategy}. We apply a dynamical mask that occludes segments outside the annotated ground-truth videos of hard samples to reduce the difficulty of hard samples.
%     }
%     \label{fig:framework} % 为图片设置引用标签
% \end{figure*}
Boundary Reflection identifies the partially annotated samples in video temporal grounding datasets.  Specifically, it uses a pretrained MLLM (\eg, Qwen2.5-VL 7B~\cite{bai2025qwen2}) to estimate the total duration of query-relevant segments outside the annotated video $\mathcal{V}_{o}=\mathcal{V}-\mathcal{V}_{a}$, where $\mathcal{V}_{a}$ denotes the segments corresponding to the GT timestamps $\mathcal{T}$. This is formulated as:
\begin{equation}
BR =  B_{\text{MLLM}}(\mathcal{P}_{BR}, \mathcal{V}_{o}),
\end{equation}
\noindent where $BR$ denotes the \emph{Boundary Reflection score}, with higher values indicating more unannotated ground-truths in the video $\mathcal{V}$. $B_{\text{MLLM}}$ represents a pretrained MLLM. $\mathcal{P}_{BR}$ denotes the boundary reflection prompt, which is sample-specific. 
Since different annotated GT segments $\mathcal{V}_{a}$ can have very different durations, the same absolute $BR$ may have a larger effect on shorter annotated intervals than on longer ones. To account for this length sensitivity, we normalize $BR$ by the duration of the annotated GT segment. Formally, we define
\begin{equation}
BR_{\mathrm{norm}} \;=\; \frac{BR}{\lvert \mathcal{V}_{a} \rvert},
\end{equation}
\noindent where $\lvert \mathcal{V}_{a} \rvert$ denotes the total duration of the annotated segments. 
\begin{figure*}[t] % [h] 代表尽量放在当前位置
    \centering
    \includegraphics[width=1.0\textwidth]{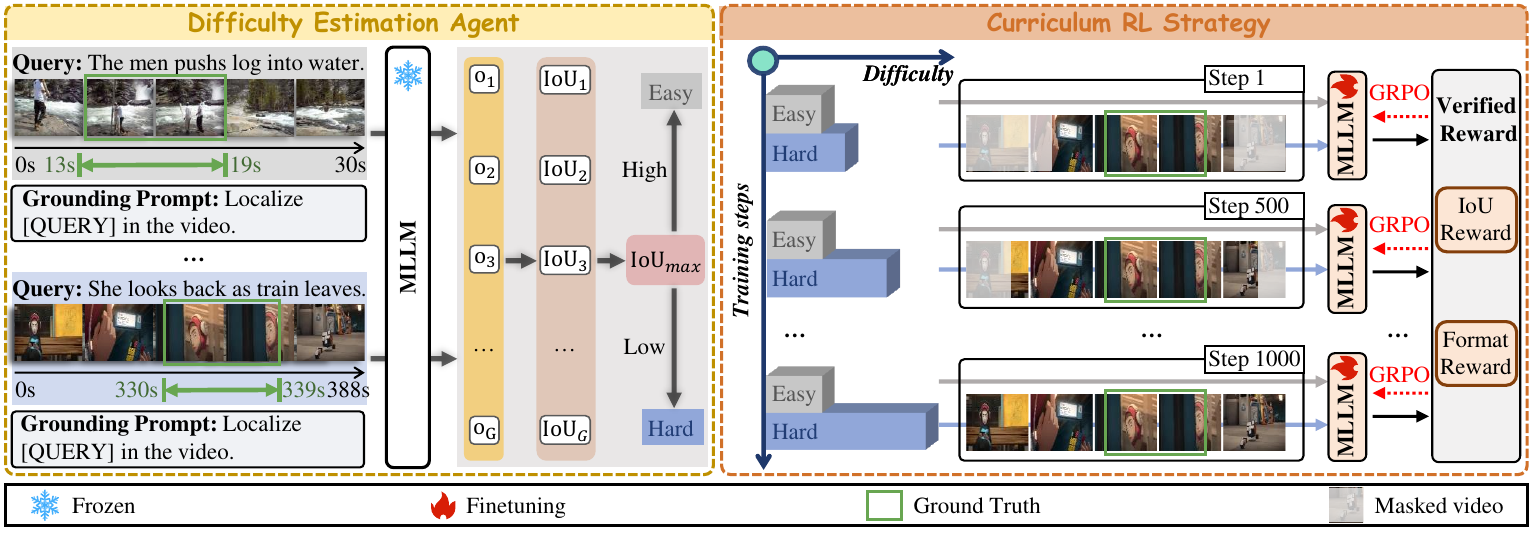} % 调整图片宽度
    \vspace{-4mm}
    \caption{\textbf{Left: Difficulty Estimation Agent}. We perform zero-shot evaluation with an MLLM to estimate the difficulty of each sample, and then split the dataset into ``hard'' and ``easy'' samples based on their predicted IoUs. \textbf{Right: Curriculum RL strategy}. For hard-to-ground samples, we dynamically mask the segment outside the annotated video according to the training step, which eases training difficulty and provides clearer preference during the RL training process.
    }
    \label{fig:framework_cl} % 为图片设置引用标签
\end{figure*}
Simply inputting $\mathcal{V}_{o}$ into MLLM may cause confusion regarding temporal information. That is, $\mathcal{V}_{o}$  consists of two non‑contiguous video segments, one before and one after the annotated video, making the model lack knowledge of their exact split points and the duration of the removed annotated segment. To address these issues, we construct the boundary reflection prompt $\mathcal{P}_{BR}$ by incorporating explicit timestamp metadata along with the grounding instruction for $\mathcal{V}_{o}$. This timestamp metadata details the original video's total duration, the timestamp of the removed interval, and the timestamps of the two subclips. The complete prompt template is provided in the supplementary materials.
 
\subsubsection{Partially Annotated Sample Filter}

By leveraging the contextual understanding capabilities of MLLMs, the Boundary Reflection score captures subtle event boundaries that may be overlooked by annotators, especially in long or complex videos. This metric thus serves as a valuable indicator for diagnosing partially annotated samples in video temporal grounding datasets.
To filter out partially-annotated samples from the dataset $\mathcal{D}$, we first discard samples with normalized boundary reflection score $BR_{\mathrm{norm}}>\tau$.
Then, from this filtered pool, we randomly draw samples until reaching our desired dataset size. By excluding partially annotated samples, we eliminate ambiguity during training and improve the overall quality of the dataset, thereby enhancing the training efficiency and the capability of the models.

\subsection{Curriculum Reinforcement Learning}
\label{sec:CRL}

Previous works have highlighted GRPO's struggles with hard samples (those exhibiting low zero-shot performance) in discrete reward tasks like ImageQA and VideoQA~\cite{xiao2025fast,park2025deepvideo}. In these scenarios,  the model often generates identical rewards (either all correct or all incorrect). This prevents GRPO from discerning relative preferences, causing advantage estimates to collapse and thus eliminating the training signal. Similarly, in the VTG task, we observe that the generated rewards for hard-to-ground samples are consistently low and undifferentiated, lacking clear preference signals. This makes it challenging to effectively train on hard-to-ground samples with GRPO.

To address this issue, we propose a curriculum reinforcement learning framework that eases the training difficulty of hard-to-ground samples. The pipeline is shown in Fig.~\ref{fig:framework_cl}. 
First, we assess the training difficulty of each samples using a Difficulty Estimation Agent. Then, we design a curriculum RL strategy that dynamically masks the videos of hard-to-ground samples according to the training steps. This allows the model to initially train on relatively easy instances with simplified video contexts, and gradually transition to more challenging ones as training progresses.

% \paragraph{Difficulty Estimation Agent.}

\subsubsection{Difficulty Estimation Agent}

The difficulty estimation agent (DEA) partitions video temporal grounding samples into ``hard'' and ``easy'' subsets. Specifically, it leverages a pretrained MLLM to evaluate the prediction difficulty. This is formulated as:
\begin{equation}
\mathcal{T}_{\mathrm{pred}} \;=\; D_{\mathrm{MLLM}}\bigl(\mathcal{P}_{ESA},\,\mathcal{V}\bigr),
\end{equation}
where $\mathcal{P}_{ESA}$ denotes the grounding prompt corresponding to each sample. We then compute the Intersection over Union (IoU) between the predicted timestamp $\mathcal{T}_{\mathrm{pred}}$ and the ground-truth timestamp $\mathcal{T}$: 
\begin{equation}
\mathrm{IoU}\bigl(\mathcal{T}_{\mathrm{pred}},\,\mathcal{T}\bigr)
\;=\;
\frac{\bigl|\,\mathcal{T}_{\mathrm{pred}}\cap \mathcal{T}\,\bigr|}
     {\bigl|\,\mathcal{T}_{\mathrm{pred}}\cup \mathcal{T}\,\bigr|}.
\end{equation}
To estimate training difficulty for GRPO, we generate the top-$k$ predictions for each sample and take the zero-shot maximum  IoU ($\mathrm{IoU}_{max}^{zs}$) among them as the difficulty score. Low $\mathrm{IoU}_{max}^{zs}$ tighten the upper bound on GRPO’s convergence, which indicates that those samples are harder to learn. 
Consequently, we regard samples with low $\mathrm{IoU}_{max}^{zs}$ as hard-to-ground samples (\ie,  $\mathrm{IoU}_{max}^{zs} \le \beta$), and the rest as easy-to-ground samples. Based on this, we adapt the curriculum RL strategy on the hard-to-ground samples to stabilize reward estimation.

\subsubsection{Curriculum RL Strategy}
We propose a curriculum RL strategy to ease the training difficulty of hard-to-ground samples. Specifically, we mask the segments $\mathcal{V}_{o}$ outside the ground-truth segments to reduce the length of negative intra-video segments. To ensure that the newly unmasked video always contains the ground-truth video, the masked segments are positioned at the beginning and end of the original video, well outside the ground-truth interval.
% time-dependent
We introduce a dynamic mask ratio $m(t)$ to control the fraction of hard-to-ground samples drawn from $\mathcal{V}_{o}$ at training step $t$ as follows:
\begin{equation}
m(t) =
\begin{cases}
m_0 \,\displaystyle\Bigl(1 - \frac{t}{w\,T}\Bigr), & 0 \le t \le wT, \\[1em]
0, & t > wT,
\end{cases}
\end{equation}
where $T$ is the total number of training steps, $w$ is the warmup ratio, and $m_0$ is the initial mask ratio.
For easy-to-ground samples, we fix $m(t)=0$ throughout training. For hard-to-ground samples, the mask ratio decays linearly from $m_0$ to $0$ over the first $wT$ training steps, and then maintain full videos as input thereafter. The variable ratio of $m$ eases the training difficulty of hard-to-ground samples, leading to better performance. 
This approach ensures that the model is gradually exposed to increasingly difficult samples during training, enabling it to consistently generate multiple outputs with distinguishable preferences at every training stage. 

To avoid positional bias, \ie, the ground-truth segment consistently appears in the same centered location, we introduce random shifts to the unmasked region within the original video. Specifically, we first compute the remaining video length by the mask ratio $m(t)$, then randomly select a window of that size from the interval encompassing the ground truth segment. This randomness encourages robustness and prevents overfitting to specific positional patterns.

We adopt the GRPO~\cite{shao2024deepseekmath} for reinforcement learning. Following VideoChat-R1~\cite{li2025videochat}, the training rewards are a format reward and an IoU reward:
\begin{equation}
    R_{VTG} = R_{format} + R_{IoU}.
\end{equation}

\section{Experiments} 
\subsection{Datasets, Evaluation, and Baseline Models} 
For training \method, we select 10\% training data from VideoMind's Grounder dataset~\cite{liu2025videomind}, yielding 21K samples across 8 datasets, \textit{i.e.}, QVHighlights (0.5K)~\cite{lei2021detecting}, DiDeMo (3.3K)~\cite{anne2017localizing}, TACoS (0.9K)~\cite{regneri2013grounding}, QuerYD (1.9K)~\cite{oncescu2021queryd}, HiREST-MR (0.8K)~\cite{zala2023hierarchical}, HiREST-Step (0.4K)~\cite{afouras2023ht}, CosMo-Cap (8.7K)~\cite{zeng2024timesuite}, and InternVid-VTime (5.4K)~\cite{huang2024vtimellm}. We evaluate \method on both VTG and grounded VideoQA benchmarks, including: 1) Charades-STA~\cite{gao2017tall} contains 6.6K indoor human-activity videos, split into 12.4K training and 3.7K test samples. 2) ActivityNet-Captions~\cite{krishna2017dense} (ANet) contains 20K YouTube videos with 37.4K training and 17.0K test samples. 3) NExT-GQA~\cite{xiao2024can} contains 1.6K daily-life videos with 3.4K validation and 5.5K test QA pairs.  4) RexTime contains 8.1K YouTube videos and is annotated with 8.7K training, 0.9K validation, and 2.1K test QA pairs. Following VideoMind~\cite{liu2025videomind}, the grounded VideoQA benchmarks consist of two sub-tasks: VTG and keyframe-based question answering tasks.

For the VTG task, we report mean Intersection over Union (mIoU) as well as recall at IoU thresholds of 0.3, 0.5, and 0.7. For the grounded VideoQA task, we additionally report Acc (accuracy), IoP (Intersection over Prediction), Acc@IoU, and Acc@IoP. The Acc@IoU and Acc@IoP denote the percentages of test samples correctly predicted based on the respective IoU or IoP threshold.

For VTG baseline models, we consider both feature-based and MLLM-based approaches: 1)  feature-based methods: 2D-TAN~\cite{zhang2020learning}, MMN~\cite{wang2022negative}, VDI~\cite{luo2023towards},  Moment-DETR~\cite{lei2021detecting}, and UniVTG~\cite{lin2023univtg}. UniVTG adopts a unified transformer and task head to jointly train VTG, highlight detection and video summarization, thereby enabling scalable multitask learning. 2) MLLM-based methods includes: a) SFT-based methods: VideoChat~\cite{li2023videochat}, 
% Valley~\cite{luo2023valley}, 
% R$^2$-Tuning~\cite{liu2024r},
VTimeLLM~\cite{huang2024vtimellm}, TimeChat~\cite{ren2024timechat}, Momentor~\cite{qian2024momentor}, 
E.T. Chat~\cite{liu2024bench},
ChatVTG~\cite{qu2024chatvtg}, VideoChat-TPO~\cite{yan2025task}, VideoMind~\cite{liu2025videomind}. VideoMind employs a chain-of-LoRA strategy to reason video tasks step by step. b) GRPO-based method:  VideoChat-R1~\cite{li2025videochat}. 

\subsection{Implementation Details}
 We leverage Qwen2.5-VL 7B~\cite{bai2025qwen2} as the backbone of our boundary reflection agent and curriculum RL strategy. For grounded VideoQA, following VideoMind~\cite{liu2025videomind}, we employ Qwen2-VL 7B~\cite{wang2024qwen2} as the QA answerer.  Videos are sampled at 2 FPS with a maximum of 384 frames. We train \method for 2 epochs using a batch size of 16. For GRPO, the top-$k$ rollout parameter is set to 8. All experiments are conducted with 16 NVIDIA A100 GPUs. The hyperparameters  $\tau$, $\beta$, $m_0$, and $w$ are set to 0, 0.3, 0.5, and 50\%, respectively.

\begin{table}[t]
\small
\tabcolsep=4pt % 适当调整列间距，您可以根据实际效果微调
\centering
\resizebox{\columnwidth}{!}{
\begin{tabular}{l|cc|cccc}
% \toprule
\Xhline{1pt}
\textbf{Method}
 & \textbf{Size}
 & \textbf{FT}
 & \textbf{R@0.3}
 & \textbf{R@0.5}
 & \textbf{R@0.7}
 & \textbf{mIoU} \\
% \midrule
\Xhline{0.7pt}

\textcolor{gray}{Moment-DETR % ~\cite{lei2021detecting}
} & 
\textcolor{gray}{--} & 
\textcolor{gray}{\checkmark} & 
\textcolor{gray}{65.8} & 
\textcolor{gray}{52.1} & 
\textcolor{gray}{30.6} & 
\textcolor{gray}{45.5} \\

\textcolor{gray}{UniVTG % ~\cite{lin2023univtg}
} & 
\textcolor{gray}{--} & 
\textcolor{gray}{\checkmark} & 
\textcolor{gray}{70.8} & 
\textcolor{gray}{58.1} & 
\textcolor{gray}{35.6} & 
\textcolor{gray}{50.1} \\

% \textcolor{gray}{R$^2$-Tuning % ~\cite{liu2024r}
% } & 
% \textcolor{gray}{--} & 
% \textcolor{gray}{\checkmark} & 
% \textcolor{gray}{70.9} & 
% \textcolor{gray}{59.8} & 
% \textcolor{gray}{37.0} & 
% \textcolor{gray}{50.9} \\

\textcolor{gray}{VideoChat-R1 % ~\cite{li2025videochat}
} & 
\textcolor{gray}{7B} & 
\textcolor{gray}{\checkmark} & 
\textcolor{gray}{-} & 
\textcolor{gray}{71.7} & 
\textcolor{gray}{50.2} & 
\textcolor{gray}{60.8} \\

% \midrule
\Xhline{0.7pt}
VTimeLLM % ~\cite{zeng2024timesuite}
 & 13B
 &
 & 55.3
 & 34.3
 & 14.7
 & 34.6 \\
TimeChat % ~\cite{ren2024timechat}
 & 7B
 &
 & 51.5
 & 32.2
 & 13.4
 & -- \\
Momentor % ~\cite{qian2024momentor}
 & 7B
 &
 & 42.6
 & 26.6
 & 11.6
 & 28.5 \\
HawkEye % ~\cite{wang2024hawkeye}
 & 7B
 &
 & 50.6
 & 31.4
 & 14.5
 & 33.7 \\
ChatVTG % ~\cite{qu2024chatvtg}
 & 7B
 &
 & 52.7
 & 33.0
 & 15.9
 & 34.9 \\
VideoChat-TPO % ~\cite{yan2025task}
 & 7B
 &
 & 58.3
 & 40.2
 & 18.4
 & 38.1 \\
% E.T. Chat~\cite{liu2024bench}
%  & 4B
%  &
%  & 65.7
%  & 45.9
%  & 20.0
%  & 42.3 \\
 VideoMind % ~\cite{liu2025videomind}
 & 7B
 &
 & 73.5
 & 59.1
 & 31.2
 & 50.2 \\
 %  Time-R1 % \textsuperscript{*}~\cite{wang2025time}
 % & 7B
 % &
 % & 78.1
 % & 60.8
 % & 35.3
 % & - \\
% \midrule
\Xhline{0.7pt}

\textbf{\method} (Ours)
 & 7B
 &
 & \textbf{79.5}
 & \textbf{64.8}
 & \textbf{39.2}
 & \textbf{55.9} \\
% \bottomrule
\Xhline{1pt}
\end{tabular}
}
\caption{Zero-shot Video Temporal Grounding on Charades-STA~\cite{gao2017tall}. \textbf{FT} indicates whether fine-tuned on the downstream training set. 
% \textsuperscript{*} means the model is trained with Gemini-1.5-Pro re-annotated data.
}
\label{tab:charades_sota}
\end{table}

\begin{table}[t]
\small
\tabcolsep=4pt % 适当调整列间距，您可以根据实际效果微调
\centering
\resizebox{\columnwidth}{!}{
\begin{tabular}{l|cc|cccc}
% \toprule
\Xhline{1pt}
\textbf{Method} & \textbf{Size} & \textbf{FT} & \textbf{R@0.3} & \textbf{R@0.5} & \textbf{R@0.7} & \textbf{mIoU} \\
% \midrule
\Xhline{0.7pt}

\textcolor{gray}{2D-TAN % ~\cite{zhang2020learning}
} & 
\textcolor{gray}{--} & 
\textcolor{gray}{\checkmark} & 
\textcolor{gray}{60.4} & 
\textcolor{gray}{43.4} & 
\textcolor{gray}{25.0} & 
\textcolor{gray}{42.5} \\

\textcolor{gray}{MMN % ~\cite{wang2022negative}
} & 
\textcolor{gray}{--} & 
\textcolor{gray}{\checkmark} & 
\textcolor{gray}{64.5} & 
\textcolor{gray}{48.2} & 
\textcolor{gray}{29.4} & 
\textcolor{gray}{46.6} \\

\textcolor{gray}{VDI % ~\cite{luo2023towards}
} & 
\textcolor{gray}{--} & 
\textcolor{gray}{\checkmark} & 
\textcolor{gray}{--} & 
\textcolor{gray}{48.1} & 
\textcolor{gray}{28.8} & 
\textcolor{gray}{--} \\
% \midrule
\Xhline{0.7pt}
VideoChat % ~\cite{li2023videochat} 
& 7B && 8.8 & 3.7 & 1.5 & 7.2 \\
% Video-LLaMA % ~\cite{zhang2023video} 
% & 7B && 6.9 & 2.1 & 0.8 & 6.5 \\
% Video-ChatGPT % ~\cite{maaz2023video} 
% & 7B && 26.4 & 13.6 & 6.1 & 18.9 \\
% Valley % ~\cite{luo2023valley} 
% & 7B && 30.6 & 13.7 & 8.1 & 21.9 \\
ChatVTG % ~\cite{qu2024chatvtg} 
& 7B && 40.7 & 22.5 & 9.4 & 27.2 \\
Momentor % ~\cite{qian2024momentor}
& 7B && 42.9 & 23.0 & 12.4 & 29.3 \\
E.T. Chat % ~\cite{liu2024bench} 
& 4B && 24.1 & 12.8 & 6.1 & 18.9 \\
VideoMind % ~\cite{liu2025videomind} 
& 7B && 48.4 & 30.3 & 15.7 & 33.3 \\
VideoChat-R1 % ~\cite{li2025videochat} 
& 7B && - & 33.4 & 17.7 & 36.6 \\
% Time-R1 % ~\cite{wang2025time} 
% & 7B && 58.6 & \textbf{39.0} & 21.4 & - \\
% \midrule
\Xhline{0.7pt}

\textbf{\method} (Ours) & 7B && \textbf{59.0} & \textbf{38.6} & \textbf{21.7} & \textbf{40.6} \\
% \bottomrule
\Xhline{1pt}
\end{tabular}
}
\caption{Zero-shot Video Temporal Grounding on ActivityNet-Captions~\cite{krishna2017dense}.}
\label{tab:anet_sota}
\end{table}

\begin{table}[t]
\small
\tabcolsep=4pt % 适当调整列间距，您可以根据实际效果微调
\centering
\resizebox{\columnwidth}{!}{
\begin{tabular}{l|cc|ccc|cc}
% \toprule
\Xhline{1pt}
\textbf{Method}
 & \textbf{Size}
 & \textbf{FT}
 & \textbf{R@0.3}
 & \textbf{R@0.5}
 & \textbf{mIoU}
 & \textbf{Acc} & \textbf{\makecell{Acc\\@IoU=0.5}} \\ % Removed Acc@IoU to match the common structure of Table 2, which has 4 metrics in the last block
% \midrule
\Xhline{0.7pt}

\textcolor{gray}{VTimeLLM % ~\cite{huang2024vtimellm}
} &
\textcolor{gray}{7B} &
\textcolor{gray}{\checkmark} &
\textcolor{gray}{43.69} &
\textcolor{gray}{26.13} &
\textcolor{gray}{29.92} &
\textcolor{gray}{57.58} &
\textcolor{gray}{17.13} \\

\textcolor{gray}{TimeChat % ~\cite{ren2024timechat}
} &
\textcolor{gray}{7B} &
\textcolor{gray}{\checkmark} &
\textcolor{gray}{40.13} &
\textcolor{gray}{21.42} &
\textcolor{gray}{26.29} &
\textcolor{gray}{49.46} &
\textcolor{gray}{10.92}\\
% \midrule
\Xhline{0.7pt}
VTimeLLM % ~\cite{huang2024vtimellm}
&
7B &
& 28.84 &
17.41 &
20.14 &
36.16 &
-- \\
TimeChat % ~\cite{ren2024timechat} 
&
7B &
& 14.42 &
7.61 &
11.65 &
40.04 &
--\\
LITA % ~\cite{huang2024lita}
&
13B &
& 29.49 &
16.29 &
21.49 &
34.44 &
-- \\

VideoMind % ~\cite{liu2025videomind}
&
7B &
& 38.22 &
25.52 &
27.61 &
74.59 &
20.20 \\
% \midrule
\Xhline{0.7pt}
% \rowcolor{gray!15}
 \textbf{\method} (Ours) &
7B &
& \textbf{41.21} &
\textbf{31.73} &
\textbf{32.24} &
\textbf{75.71} &
\textbf{25.86}\\
% \bottomrule
\Xhline{1pt}
\end{tabular}
}
\caption{Zero-shot grounded VideoQA on ReXTime~\cite{chen2024rextime}. }
\label{tab:rextime_sota}
\end{table}

\begin{table}[h]
\small
\renewcommand{\arraystretch}{1.2}  % 数值 >1 会增大行高
\tabcolsep=4pt % 调整列间距
\centering
\resizebox{\columnwidth}{!}{%
\begin{tabular}{l|c|
            @{\hskip 2pt}c@{\hskip 2pt}
            c@{\hskip 2pt}
            c@{\hskip 2pt}|
            c|c|c|c}
\Xhline{1pt}
\multirow{2}{*}{\textbf{Method}} 
  & \multirow{2}{*}{\textbf{Size}} 
  & \multicolumn{3}{c|}{\textbf{IoU}} 
  & \multirow{2}{*}{\textbf{Acc}}
  & \multirow{2}{*}{\textbf{\makecell{Acc\\{\footnotesize@IoU=0.5}}}}
  & \multirow{2}{*}{\textbf{mIoP}} 
  & \multirow{2}{*}{\textbf{\makecell{Acc\\{\footnotesize@IoP=0.5}}}} \\
\cline{3-5}
  & 
  & {\footnotesize R@0.3} & {\footnotesize R@0.5} & {\footnotesize mIoU} 
  & 
  & 
  & 
  &  \\ 
\Xhline{0.5pt}
FrozenBiLM NG+  % \cite{yang2022zero}     
& 890M & 13.5 &  6.1 &  9.6 & 70.8 &  –   & 24.2 & 17.5 \\
SeViLA %  \cite{yu2023self}        
& 4B   & 29.2 & 13.8 & 21.7 & 68.1 &  –   & 29.5 & 16.6 \\
LangRepo %  \cite{kahatapitiya2024language} 
& 8$\times$7B & -- & 12.2 & 18.5 & –    &  –   & 31.3 & 17.1 \\
LLoVi  % \cite{zhang2023simple}       
& 1.8T & --   & 15.3 & 20.0 & –    &  –   & 37.3 & 24.3 \\
HawkEye %  \cite{wang2024hawkeye}     
& 7B   & 37.0 & 19.5 & 25.7 & –    &  –   & --   & --   \\
VideoChat-TPO %  \cite{yan2025task}        
& 7B   & 41.2 & 23.4 & 27.7 & –    &  –   & 35.6 & 25.5 \\
\textcolor{gray}{VideoMind} %  \cite{liu2025videomind}      
& \textcolor{gray}{7B}   & \textcolor{gray}{50.2} & \textcolor{gray}{25.8} & \textcolor{gray}{31.4}
  & \textcolor{gray}{76.9}$^{\dagger}$ & \textcolor{gray}{20.2}$^{\dagger}$ & \textcolor{gray}{39.0} & \textcolor{gray}{28.2} \\

\Xhline{0.5pt}
\textbf{\method} (Ours)                   
& 7B   & \textbf{50.6} & \textbf{28.0} & \textbf{34.2} 
  & \textbf{77.4} & \textbf{22.5} & \textbf{38.3} & \textbf{26.5} \\
\Xhline{1pt}
\end{tabular}%
}
\caption{Zero-shot grounded VideoQA on NExT-GQA \cite{xiao2024can}. Here, Acc@IoP=0.5 is also termed Acc@GQA. $^{\dagger}$ indicates our evaluation using the official checkpoint. Note that VideoMind's training data includes the NExT-GQA training set, so its performance is not strictly zero-shot.}
\label{tab:nextgqa_sota}
\end{table}

\subsection{Comparison with State-of-the-Art}

As shown in Table \ref{tab:charades_sota}, \ref{tab:anet_sota}, \ref{tab:rextime_sota}, and \ref{tab:nextgqa_sota}, \method achieves state-of-the-art performance across these four benchmarks. Specifically, on Charades-STA and ANet, \method outperforms previous SoTA by margins of 5.7\% and 4.0\% on mIoU, respectively. 
For the grounded VideoQA task, \method outperforms VideoMind~\cite{liu2025videomind}  on RexTime by 4.63\%  in mIoU and 5.66\% in Acc@IoU@0.5, and \method outperforms VideoChat-TPO~\cite{yan2025task}  on NExT-GQA by 6.5\% in mIoU and 1.0\% in Acc@IoP@0.5. 

 Overall, these results highlight that the grounding timestamps generated by our \method are of higher quality.

\begin{table}[t]
   \footnotesize
  \tabcolsep=2pt
  \renewcommand{\arraystretch}{1.05}
  \centering
  \resizebox{1\linewidth}{!}{%
    \begin{tabular}{c |c c c | c | c c | c c}
      \Xhline{1pt}
      \multirow{2}{*}{\makecell{Training \\Mode}} &
      \multirow{2}{*}{\makecell{Sample\\Ratio}} &
      \multirow{2}{*}{\makecell{Sample\\Method}} &
      \multirow{2}{*}{\makecell{DM}} &
      \multirow{2}{*}{\makecell{Normalized \\Cost}} &
      \multicolumn{2}{c|}{Charades-STA} &
      \multicolumn{2}{c}{ANet} \\
      \cline{6-7} \cline{8-9}
      & & & & & R@0.5 & mIoU & R@0.5 & mIoU \\
      \Xhline{0.7pt}
      \multirow{4}{*}{SFT}
      & 100\% & Random    & $\times$     & 1.000 & 40.1       & 38.3   & 21.7    & 26.4  \\
      & 10\%  & Random    & $\times$     & \textbf{0.100}  & 32.0       & 31.2   & 20.0    & 25.0  \\
      & 10\%  & BRA       & $\times$     & 0.139  & 43.5       & 39.4   & 26.5    & 30.2  \\
      & 10\%  & BRA       & $\checkmark$ & 0.205  & \textbf{44.5} & \textbf{40.5} & \textbf{29.0} & \textbf{32.5} \\
      \Xhline{0.5pt}
      \multirow{4}{*}{GRPO}
      & 100\% & Random    & $\times$     & 1.000 & 64.5 & 55.4   & 36.1    & 38.1  \\
      & 10\%  & Random    & $\times$     & \textbf{0.100}  & 62.9       & 53.8   & 35.1    & 37.4  \\
      & 10\%  & BRA       & $\times$     & 0.127  & 64.3       & 55.3   & 37.6    & 39.8  \\
      & 10\%  & BRA       & $\checkmark$ & 0.174  & \textbf{64.8}       & \textbf{55.9} & \textbf{38.6} & \textbf{40.6} \\
      \Xhline{1pt}
    \end{tabular}%
  }
  \caption{Ablations of the boundary reflection agent (BRA) and dynamic masking (DM) under both GRPO and SFT. 
  Normalized cost is expressed in relative A100 GPU-hours, including preprocessing and training. 
  }
  \label{tab:ablation_combined}
\end{table}

\begin{table}[t]
  \footnotesize
  \tabcolsep=2pt
  \renewcommand{\arraystretch}{1.05}
  \centering
  \resizebox{1\linewidth}{!}{%
    \begin{tabular}{c c c | c c | c c}
      \Xhline{1pt}
      Sample & Sample & Normalized & \multicolumn{2}{c|}{Charades-STA} & \multicolumn{2}{c}{ANet} \\
      \cline{4-5} \cline{6-7}
      Ratio & Method & Cost & R@0.5 & mIoU & R@0.5 & mIoU \\
      \Xhline{0.7pt}
      10\%  & Random    & \bf 0.100 & 62.9 & 53.8 & 35.1 & 37.4 \\
      10\%  & COINCIDE~\cite{lee2024concept}  & 0.879 & 63.3 & 54.1 & 35.9 & 37.7 \\
      10\%  & PreSel~\cite{safaei2025filter}    & 0.187 & 63.5 & 54.4 & 36.2 & 38.3 \\
      10\%  & BRA (GT)   & 0.119 & 62.1 & 53.1 & 34.1 & 36.8 \\
      10\%  & BRA    & 0.127 & \bf 64.3 & \bf  55.3 & \bf  37.6 & \bf  39.8 \\
      \Xhline{1pt}
    \end{tabular}%
  }
  \caption{Ablation of sampling methods under GRPO. ``BRA (GT)'' indicates boundary reflection applied to the alignment between the query and the ground-truth interval.
  % Normalized cost is expressed in relative A100 GPU-hours, including preprocessing and training. 
  }
  \label{tab:ablation_data_selection}
\end{table}

\begin{table}[t]
  \footnotesize
  \centering

  \setlength{\tabcolsep}{2pt}
  \renewcommand{\arraystretch}{1.05}

\resizebox{1\linewidth}{!}{%
  \begin{tabular}{c|c|c|c|cc|cc}
    \Xhline{1pt}
    \multirow{2}{*}{Setting}
      & \multirow{2}{*}{\makecell{Thre.\\$\tau$}}
      & \multirow{2}{*}{ADT}
      & \multirow{2}{*}{ASR}
      & \multicolumn{2}{c|}{Charades-STA}
      & \multicolumn{2}{c}{ANet} \\
    \cline{5-8}
    & & & & R@0.5 & mIoU & R@0.5 & mIoU \\
    \Xhline{0.7pt}
    base
      & 0
      & Hard
      & 50\%
      & \textbf{64.8} & \textbf{55.9}
      & \textbf{38.6} & \textbf{40.6} \\
    \Xhline{0.5pt}
    \multirow{3}{*}{(A)}
      & \cellcolor{gray!20}0.1  & Hard & 50\%  & 64.5 & 55.4 & 38.1 & 39.9 \\
      & \cellcolor{gray!20}1    & Hard & 50\%  & 64.3 & 55.1 & 37.4 & 39.1 \\
      & \cellcolor{gray!20}10   & Hard & 50\%  & 63.7 & 54.7 & 36.1 & 38.2 \\
    \Xhline{0.5pt}
    \multirow{2}{*}{(B)}
      & 0 & \cellcolor{gray!20}None  & 50\% & 64.3 & 55.3 & 37.6 & 39.8 \\
      & 0 & \cellcolor{gray!20}All   & 50\% & 63.7 & 54.1 & 36.5 & 38.0 \\
    \Xhline{0.5pt}
    \multirow{2}{*}{(C)}
      & 0 & Hard & \cellcolor{gray!20}30\%  & 64.6 & 55.7 & 38.2 & 40.3 \\
      & 0 & Hard & \cellcolor{gray!20}100\%& 64.1 & 54.3 & 37.4 & 39.6 \\
    \Xhline{1pt}
  \end{tabular}
}
  \caption{Ablations for (A) threshold $\tau$ of $BR_{\mathrm{norm}}$, (B) augmentation data type (ADT), and (C) augmentation step ratio (ASR).}
  \label{tab:ablation_hyper_all}
\end{table}

\begin{figure}[t] % [h] 代表尽量放在当前位置
    \centering
    \includegraphics[width=1.0\linewidth]{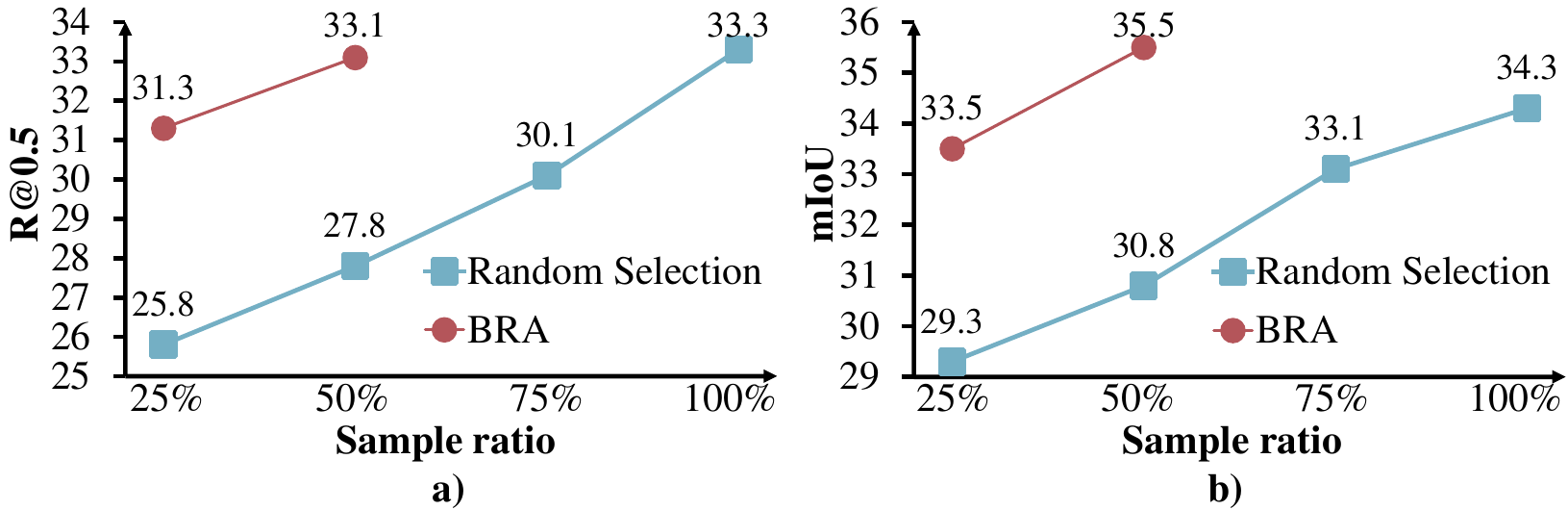} % 调整图片宽度
    \vspace{-4mm}
    \caption{Out-of-domain evaluation across varying sample ratios. Models are trained with GRPO on the Charades-STA dataset, and tested on the ActivityNet-Captions dataset. Results are presented for a) R@0.5 and b) mIoU metrics.
    }
    \label{fig:cross_dataset_eval} % 为图片设置引用标签
\end{figure}

\subsection{Ablation Study}
We conduct comprehensive ablation experiments on  Charades-STA and ANet to explore the effects of different modules, the hyperparameters, more ablation, and more evaluation of boundary reflection agent (BRA), and the hyperparameters for curriculum reinforcement learning. 
The findings from these ablation studies provide more in-depth insights into the critical elements that contribute to the effectiveness of our proposed framework.

% \paragraph{Ablation of different modules.}
\subsubsection{Ablation of different modules}
\textbf{(1) Ablation in GRPO}. We ablate the two modules of \method, \ie, the boundary reflection agent (BRA) and the dynamic masking (DM) in our curriculum RL strategy. From the lower half of Table \ref{tab:ablation_combined}, we make the following observations: (a) With only 10\% of the data,  BRA beats random selection by 1.5\% mIoU on Charades-STA and 2.4\% mIoU on ANet. Compared to full-sample training (which runs 5.8× slower in terms of runtime), BRA still matches the performance on Charades-STA and exceeds the performance by 1.7\% on ANet.  These results confirm that unambiguous, fully annotated samples are crucial for video temporal grounding. (b) The DM further improves the performance by 0.6\% mIoU on Charades-STA and 0.8\% mIoU on ANet, which highlights the importance of reducing training difficulty for hard-to-ground samples. (c) The preprocessing cost of BRA and DM is relatively small (2.7\% for BRA, 4.7\% for DM), since evaluation is much faster than GRPO training. DM is slower than BRA because it predicts multiple top-k outputs, whereas BRA only predicts top-1.  The relatively low overhead demonstrates that \method is scalable to larger datasets.

\textbf{(2) Ablation in SFT}. To further evaluate the quality of the data curated by our method, we also integrate BRA and DM into SFT to validate their generalization. From the top half of Table \ref{tab:ablation_combined}, we observe: (a) With only 10\% of the data, BRA boosts the random baseline by 8.2\% mIoU on Charades-STA and 5.2\% mIoU on ANet. BRA also surpasses the full sample training, underscoring the data efficiency of unambiguous, fully annotated samples. (b) The DM further improves by 1.1\% mIoU on Charades-STA and 2.3\% mIoU on ANet, which also highlights the benefit of reducing training difficulty for hard-to-ground samples.  (c) The gains of BRA and DM under GRPO are smaller than under SFT, underscoring GRPO’s robustness on partially annotated and hard-to-ground samples. We attribute this to GRPO’s narrow reward range and Gaussian smoothing, which reduce sensitivity to the signals from ambiguous and challenging samples, whereas SFT’s unconstrained cross-entropy loss is more prone to such difficult samples.

\begin{figure}[t] % [h] 代表尽量放在当前位置
    \centering
    \includegraphics[width=1.0\linewidth]{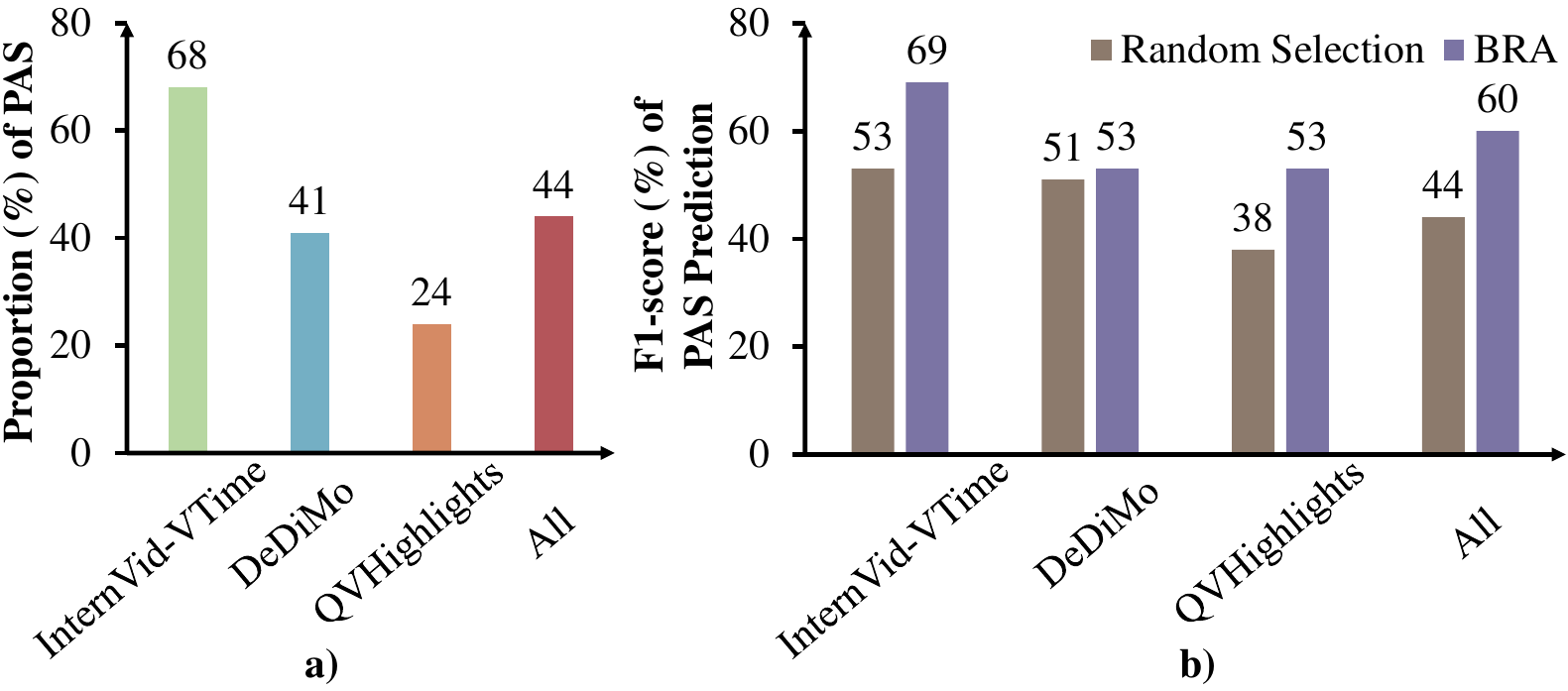} % 调整图片宽度
    \vspace{-4mm}
    \caption{
        a) Proportion of Partially Annotated Sample (PAS) in the manually labeled subset. We randomly select 100 samples from each of three datasets and annotate them. b) F1-score of PAS prediction using random selection and BRA. GT labels are drawn from the manually labeled subset.
    }
    \label{fig:BRA_quality} % 为图片设置引用标签
\end{figure}
% a) Missing Annotation ratio of the manually labeled subset. We randomly select 100 samples from each of three datasets and annotate them.  b) F1-score of missing annotation prediction using random selection and BRA. GT labels are drawn from the manually labeled subset.

% \begin{figure}[t] % [h] 代表尽量放在当前位置
%     \centering
%     \includegraphics[width=1.0\linewidth]{AnonymousSubmission/LaTeX/figs/0924_case.pdf} % 调整图片宽度
%     \vspace{-4mm}
%     \caption{Qualitative examples of the ground truth, VideoChat-R1, and our method (\method). Examples are from the ActivityNet-Captions dataset.
%     }
%     \label{fig:case_study} % 为图片设置引用标签
% \end{figure}
\begin{figure}[t] % [h] 代表尽量放在当前位置
    \centering
    \includegraphics[width=1.0\linewidth]{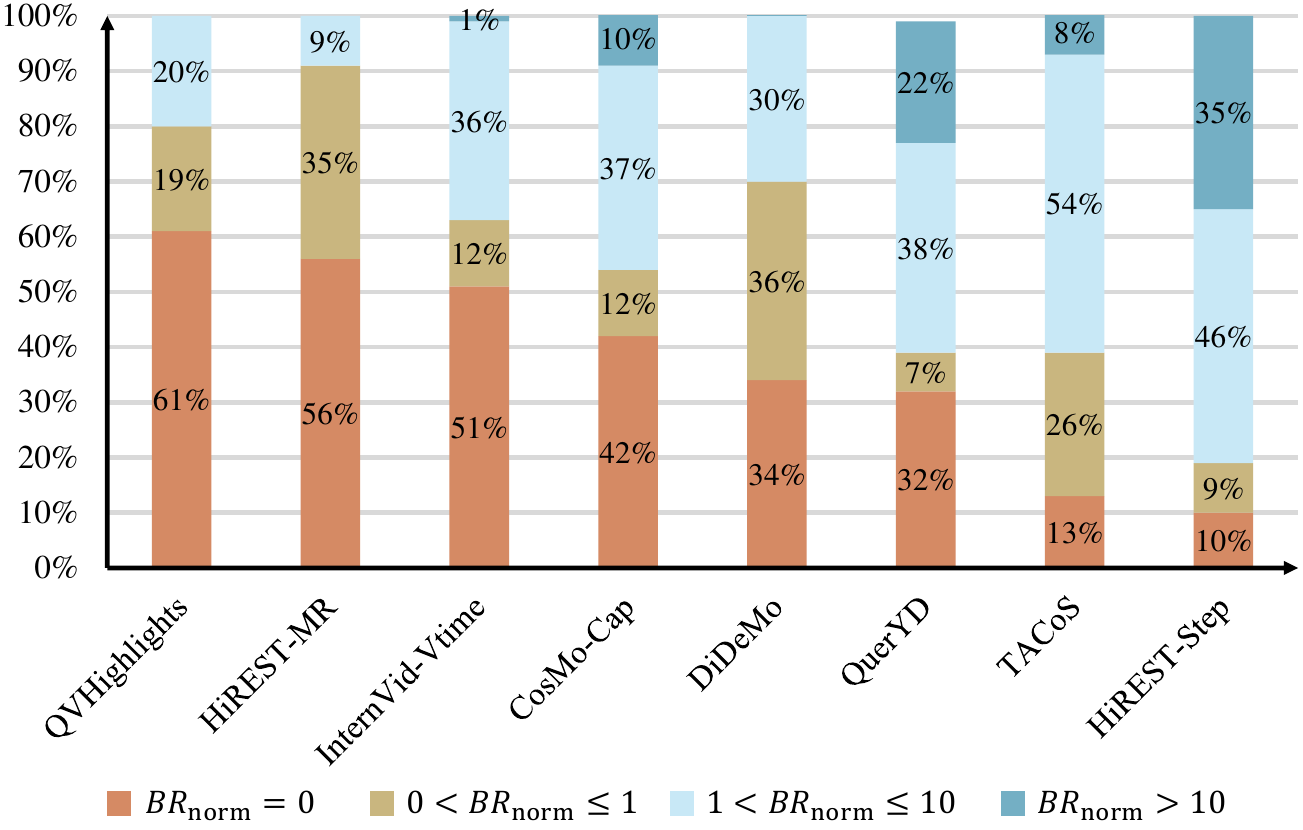} % 调整图片宽度
    \vspace{-4mm}
    \caption{Distribution of $BR_{\mathrm{norm}}$ across the eight datasets in our training set.
    }
    \label{fig:BR_norm_distribution} % 为图片设置引用标签
\end{figure}
% \begin{figure}[t] % [h] 代表尽量放在当前位置
%     \centering
%     \includegraphics[width=1.0\linewidth]{AnonymousSubmission/LaTeX/figs/0924_IoU_max_difficulty.pdf} % 调整图片宽度
%     \vspace{-4mm}
%     \caption{Training-step curves of Reward and Top-1 IoU for samples grouped by $\mathrm{IoU}_{max}^{zs}$ ranges. We observe that samples in the low $\mathrm{IoU}_{max}^{zs}$ range exhibit unstable rewards and poorer convergence (lower Top-1 IoU).
%     }
%     \label{fig:case_study} % 为图片设置引用标签
% \end{figure}

% \begin{figure}[t] % [h] 代表尽量放在当前位置
%     \centering
%     \includegraphics[width=1.0\linewidth]{figures/0804_training_curve.pdf} % 调整图片宽度
%     \vspace{-4mm}
%     \caption{Training curves of GRPO on four difficulty tiers defined by zero-shot IoU ($\mathrm{IoU}_{\mathrm{zs}}$): a) average Top-1 IoU; b) average advantages. Each point denotes the average result over 50 steps.
%     }
%     \label{fig:training_curve} % 为图片设置引用标签
% \end{figure}
\begin{figure}[t] % [h] 代表尽量放在当前位置
    \centering
    \includegraphics[width=1.0\linewidth]{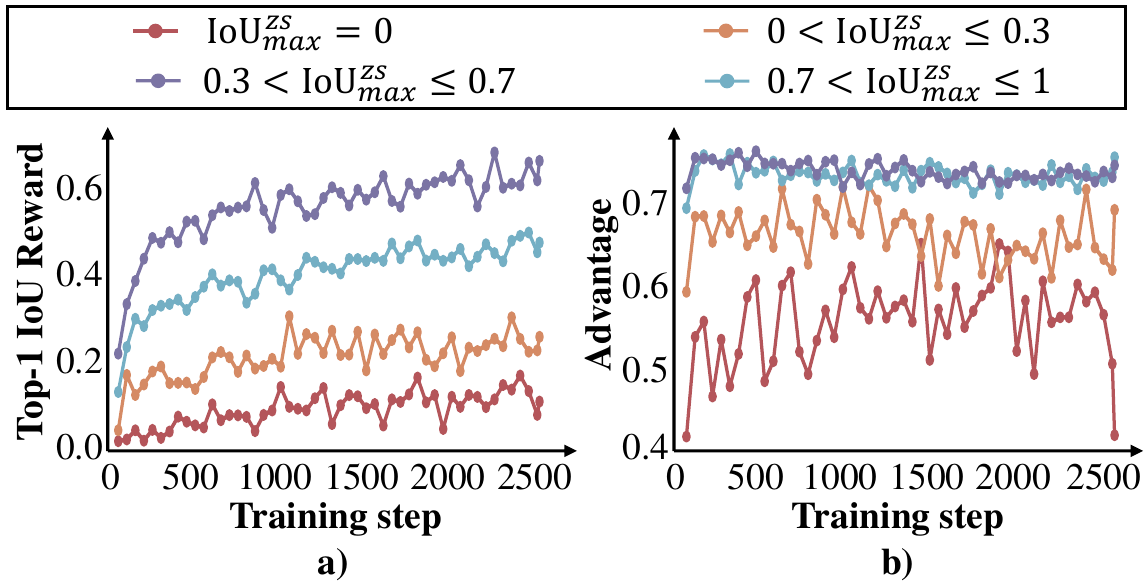} % 调整图片宽度
    \vspace{-4mm}
    \caption{a) Top-1 IoU Reward curves for different groups, categorized by their zero-shot maximum IoU  ($\mathrm{IoU}_{max}^{zs}$). b) Advantage curves of different groups, categorized by their  zero-shot maximum IoU ($\mathrm{IoU}_{max}^{zs}$). 
    }
    \label{fig:iou_max_training_difficulty} % 为图片设置引用标签
\end{figure}
% \begin{figure}[t] % [h] 代表尽量放在当前位置
%     \centering
%     \includegraphics[width=1.0\linewidth]{figures/0804_training_curve.pdf} % 调整图片宽度
%     \vspace{-4mm}
%     \caption{Training curves of GRPO on four difficulty tiers defined by zero-shot IoU ($\mathrm{IoU}_{\mathrm{zs}}$): a) average Top-1 IoU; b) average advantages. Each point denotes the average result over 50 steps.
%     }
%     \label{fig:training_curve} % 为图片设置引用标签
% \end{figure}
\begin{figure}[t] % [h] 代表尽量放在当前位置
    \centering
    \includegraphics[width=1.0\linewidth]{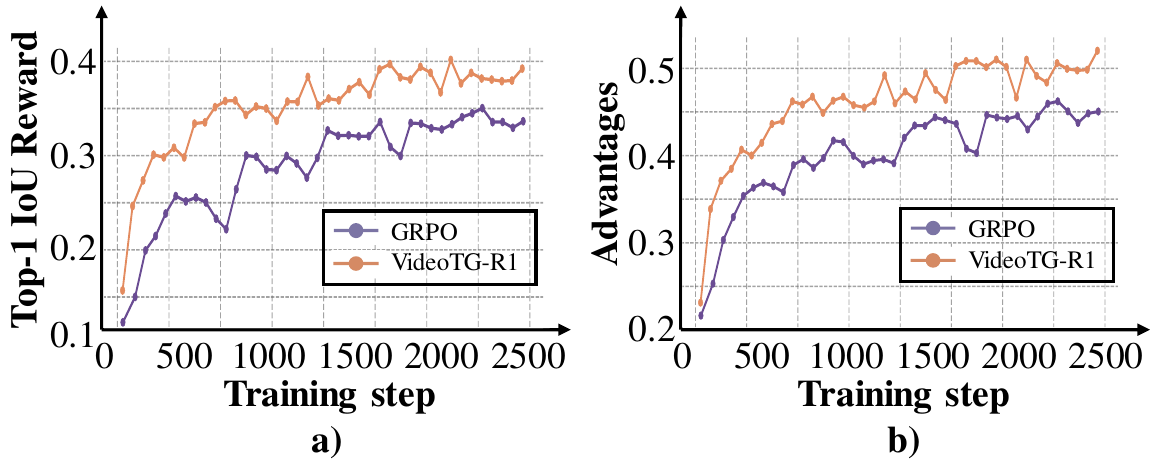} % 调整图片宽度
    \vspace{-4mm}
    \caption{ a) Top-1 IoU Reward curves of GRPO and \method (Ours). b) Advantage curves of GRPO and \method (Ours).
    }
    \label{fig:training_curves} % 为图片设置引用标签
\end{figure}
% \paragraph{Ablation of boundary reflection agent.} 
\subsubsection{Ablation of boundary reflection agent} 

\textbf{(1) Data selection strategy} 
Table~\ref{tab:ablation_data_selection} presents a comparison between various data selection strategies, including random, COINCIDE~\cite{lee2024concept}, PreSel~\cite{safaei2025filter}, BRA (GT), and our BRA. Here, BRA (GT) evaluates the alignment between the query and the ground-truth interval, whereas the default BRA evaluates whether the query event occurs inside non-GT intervals.
We make two main observations. (a) BRA achieves the best overall performance and low normalized cost. We attribute this to the stronger interpretability of BRA for VTG tasks: partially annotated samples are widespread in VTG datasets (see Fig.~\ref{fig:BRA_quality} a)), and BRA can filter out these partially annotated samples. (b) BRA (GT) performs substantially worse than BRA, and even underperforms the random baseline. This result stems from that the query events of the GT intervals are mostly human-annotated and of high quality, so additional model-based re-checking can introduce spurious noise. By contrast, non-GT intervals lack manual verification, and using the model to inspect non-GT intervals produces meaningful gains.

\textbf{(2) Hyperparameters}. 
We ablate the threshold  $\tau$ of $BR_{\mathrm{norm}}$ in the boundary reflection agent, \ie, we discard samples whose normalized boundary reflection score $BR_{\mathrm{norm}}>\tau$.  $\tau$ controls the noise tolerance for samples with annotation omission. From Table \ref{tab:ablation_hyper_all} (A), we can observe that raising the threshold $\tau$ admits more ambiguous, partially annotated samples and steadily degrades performance, highlighting the high quality of fully-annotated samples.

\textbf{(3) Generalization for out-of-domain evaluation}.
Following VideoChat-R1~\cite{li2025videochat}, we also apply BRA to out-of-domain scenarios, \ie, training on Charades-STA and evaluating on ANet, to further validate its effectiveness.
Fig.\ref{fig:cross_dataset_eval} illustrates the out-of-domain performance under varying sample ratios. We observe that: a) At 25\% and 50\% sample ratios, BRA significantly outperforms random selection, exceeding it by 5.5\% and 5.3\% in R@0.5, and by 4.2\% and 4.7\% in mIoU, respectively. These results demonstrate the superior generalization of BRA. b) Notably, BRA at 50\% matches the performance of training with 100\% of the samples, highlighting the data efficiency of unambiguous, fully annotated samples. c) As shown by the red line in Fig.~\ref{fig:cross_dataset_eval}, using a larger amount of high-quality data curated by BRA consistently improves model performance. This further validates the effectiveness of our approach and highlights the importance of high-quality data.

\textbf{(4) Label quality of partially annotated sample prediction}.
To evaluate the label quality of BRA for partially annotated samples, we manually label 100 samples from each of three datasets: InternVid-VTime~\cite{huang2024vtimellm}, DeDiMo~\cite{anne2017localizing}, and QVHighlights~\cite{lei2021detecting}. Then we compute the F1-score of random selection and our BRA to evaluate their selection quality. From Fig.\ref{fig:BRA_quality} a), it is evident that partially annotated samples are a prevalent phenomenon. From Fig.~\ref{fig:BRA_quality} b), we can observe that  BRA outperforms random selection across every dataset, especially on the model-annotated InternVid-VTime, indicating that our BRA can effectively identify partially annotated samples. 

\subsubsection{Ablation of curriculum RL strategy}
We systematically evaluate two pivotal hyperparameters of curriculum RL strategy: i) Augmentation data type: which types of samples should be masked. ii) Augmentation step ratio: the proportion of steps during which masking is applied.
From Table \ref{tab:ablation_hyper_all} (B) and (C), we find: a) Masking all samples hurts performance. We attribute this to overfitting: excessively masking easy-to-ground samples makes them ``too easy", encouraging the model to memorize superficial artifacts rather than learn robust alignment. 
b) Going to 100\% masking of hard-to-ground samples also slightly degrades results, likely because excessive masking creates boundary artifacts that widen the training/test distribution gap. Including some original hard-to-ground samples is thus crucial for exposing the model to realistic video.

\subsubsection{Distribution of Boundary Reflection Score}
% \subsection{}
To demonstrate the generality of partially annotated samples (PAS) and their variation across datasets, we present the model-predicted 
$BR_{\mathrm{norm}}$ distributions on the training sets of multiple datasets. $BR_{\mathrm{norm}}$ denotes the ratio between the duration of ground-truth that is missing (\ie, unannotated) to the duration of the originally annotated ground-truth. From Fig.\ref{fig:BR_norm_distribution}, we observe that: 1) Across all datasets, the proportion of PAS segments is substantial ($\ge 39\%$), which is consistent with the results in Fig.~\ref{fig:BRA_quality} a), \ie, partially annotated samples are a prevalent phenomenon. 
2) The HiREST-Step dataset shows the highest PAS proportion, which we attribute to its queries consisting of atomic, single-step events that frequently occur in the video.
3) The QVHighlights dataset exhibits the lowest PAS proportion, since its annotators are instructed to annotate multiple occurrences of a query in the video. However, as shown in Fig.~\ref{fig:BRA_quality} a), there remains a non-negligible PAS proportion after careful manual inspection.

\subsubsection{Training Difficulty of Samples with Low Zero-shot IoU}

To verify that our proposed zero-shot maximum IoU ($\mathrm{IoU}_{max}^{zs}$) metric can discriminate the training difficulty of GRPO across samples, we partition the full training set into four groups according to each sample’s $\mathrm{IoU}_{max}^{zs}$. The four groups correspond to the following intervals:
\[
[0,0],\quad (0,\,0.3],\quad (0.3,\,0.7],\quad (0.7,\,1].
\]

Their proportions are 14.8\%, 15.8\%, 37.4\%, and 32.0\%, respectively. For each group, we track the average Top-1 IoU reward and advantage as a function of training step, and report the curves in Fig.~\ref{fig:iou_max_training_difficulty}. The plots reveal two clear trends: 1) Samples with low $\mathrm{IoU}_{max}^{zs}$
 achieve substantially lower Top-1 IoU reward throughout training, \ie, they are harder to learn and behave as hard samples; 2) Low $\mathrm{IoU}_{max}^{zs}$
 groups also exhibit lower and more unstable advantage trajectories, indicating that they are more difficult for GRPO to learn. Overall, these results validate the effectiveness of the metric 
$\mathrm{IoU}_{max}^{zs}$ for distinguishing training difficulty of GRPO across samples.

\subsubsection{Training Curves of GRPO and \method}

We compare the training curves of GRPO and our VideoTG-R1 in Fig.~\ref{fig:training_curves}, including Top-1 IoU reward curves and advantage curves. The plots show that VideoTG-R1 consistently outperforms GRPO: it rises faster in the early training steps, achieves a higher Top-1 IoU reward throughout training, and also attains higher advantage values across the whole run. This consistent gap indicates that VideoTG-R1 both speeds up learning and improves localization quality, producing more accurate and stable grounding than the naive GRPO.

\begin{figure}[t] % [h] 代表尽量放在当前位置
    \centering
    \includegraphics[width=1.0\linewidth]{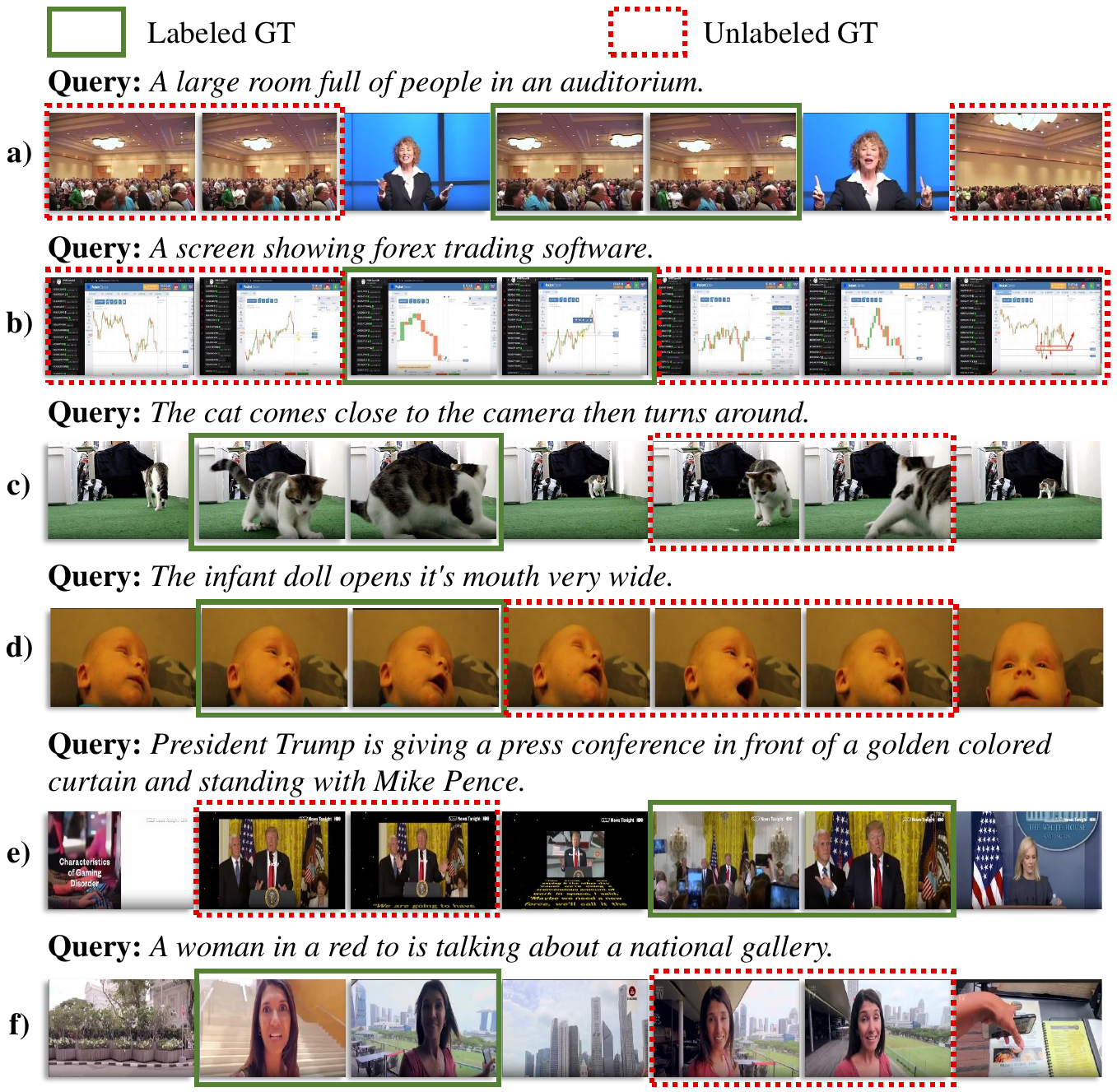} % 调整图片宽度
    \vspace{-4mm}
    \caption{More partially annotated examples. Examples (a) and (b) are from the InternVid-VTime dataset, (c) and (d) are from the DiDeMo dataset, (e) and (f) are from the QVHighlights dataset. 
    }
    \label{fig:multiple_gt_case} % 为图片设置引用标签
\end{figure}

\subsubsection{More partially annotated samples}
To demonstrate the prevalence of partially annotated samples in existing VTG datasets,  we present a series of representative examples in Fig.~\ref{fig:multiple_gt_case}. Specifically, we categorize these partially annotated samples into three types: (1) Monotonous video content without clear scene transitions (Fig.~\ref{fig:multiple_gt_case} a), b), and d)).
(2) Clips with back-and-forth switching (Fig.~\ref{fig:multiple_gt_case} e), f)).
(3) Sequences with repeated motion patterns (Fig.~\ref{fig:multiple_gt_case} c)). 
These examples, taken from the InternVid-VTime, DiDeMo, and QVHighlights datasets, underscore the inherent ambiguity in the existing VTG dataset.

% \paragraph{Qualitative Analysis.}
\subsubsection{Qualitative Analysis}
\begin{figure}[t] % [h] 代表尽量放在当前位置
    \centering
    \includegraphics[width=1.0\linewidth]{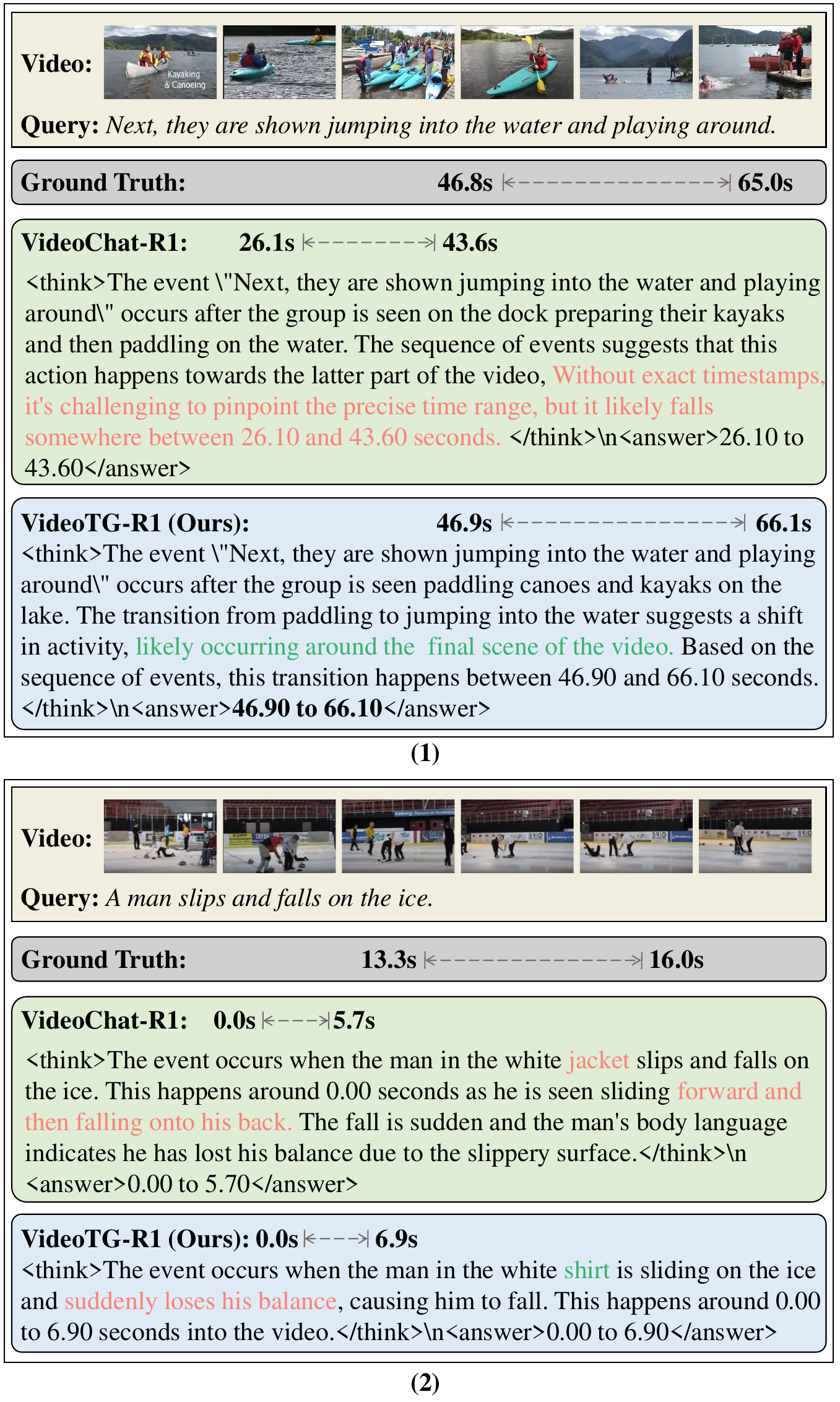} % 调整图片宽度
    \vspace{-4mm}
    \caption{Qualitative examples of the ground truth, VideoChat-R1, and \method (Ours). Examples are from the ActivityNet-Captions dataset.
    }
    \label{fig:case_study} % 为图片设置引用标签
\end{figure}
% \begin{figure}[t] % [h] 代表尽量放在当前位置
%     \centering
%     \includegraphics[width=1.0\linewidth]{AnonymousSubmission/LaTeX/figs/0924_br_norm_distribution.pdf} % 调整图片宽度
%     \vspace{-4mm}
%     \caption{Distribution of $BR_{\mathrm{norm}}$ across the eight datasets in our training set.
%     }
%     \label{fig:case_study} % 为图片设置引用标签
% \end{figure}
% \begin{figure}[t] % [h] 代表尽量放在当前位置
%     \centering
%     \includegraphics[width=1.0\linewidth]{AnonymousSubmission/LaTeX/figs/0924_IoU_max_difficulty.pdf} % 调整图片宽度
%     \vspace{-4mm}
%     \caption{Training-step curves of Reward and Top-1 IoU for samples grouped by $\mathrm{IoU}_{max}^{zs}$ ranges. We observe that samples in the low $\mathrm{IoU}_{max}^{zs}$ range exhibit unstable rewards and poorer convergence (lower Top-1 IoU).
%     }
%     \label{fig:case_study} % 为图片设置引用标签
% \end{figure}

Fig.~\ref{fig:case_study} shows the qualitative comparison between a baseline VideoChat-R1 and our \method. From these case studies, we draw the following observations: (1) In Fig.~\ref{fig:case_study} a), our method’s thinking process more effectively supports grounding reasoning and produces superior results compared to the baseline, demonstrating that \method generates more accurate query-relevant temporal locations. 
(2) In Fig.~\ref{fig:case_study} b),  both the baseline and our method fail to accurately localize the query due to the video content of rapid movements. However, the baseline exhibits more hallucinations (\ie, the man is wearing a white shirt, yet VideoChat-R1 erroneously predicts a jacket), highlighting that fully annotated samples can drive the model toward better optimization directions.

\begin{figure}[t] % [h] 代表尽量放在当前位置
    \centering
    \includegraphics[width=1.0\linewidth]{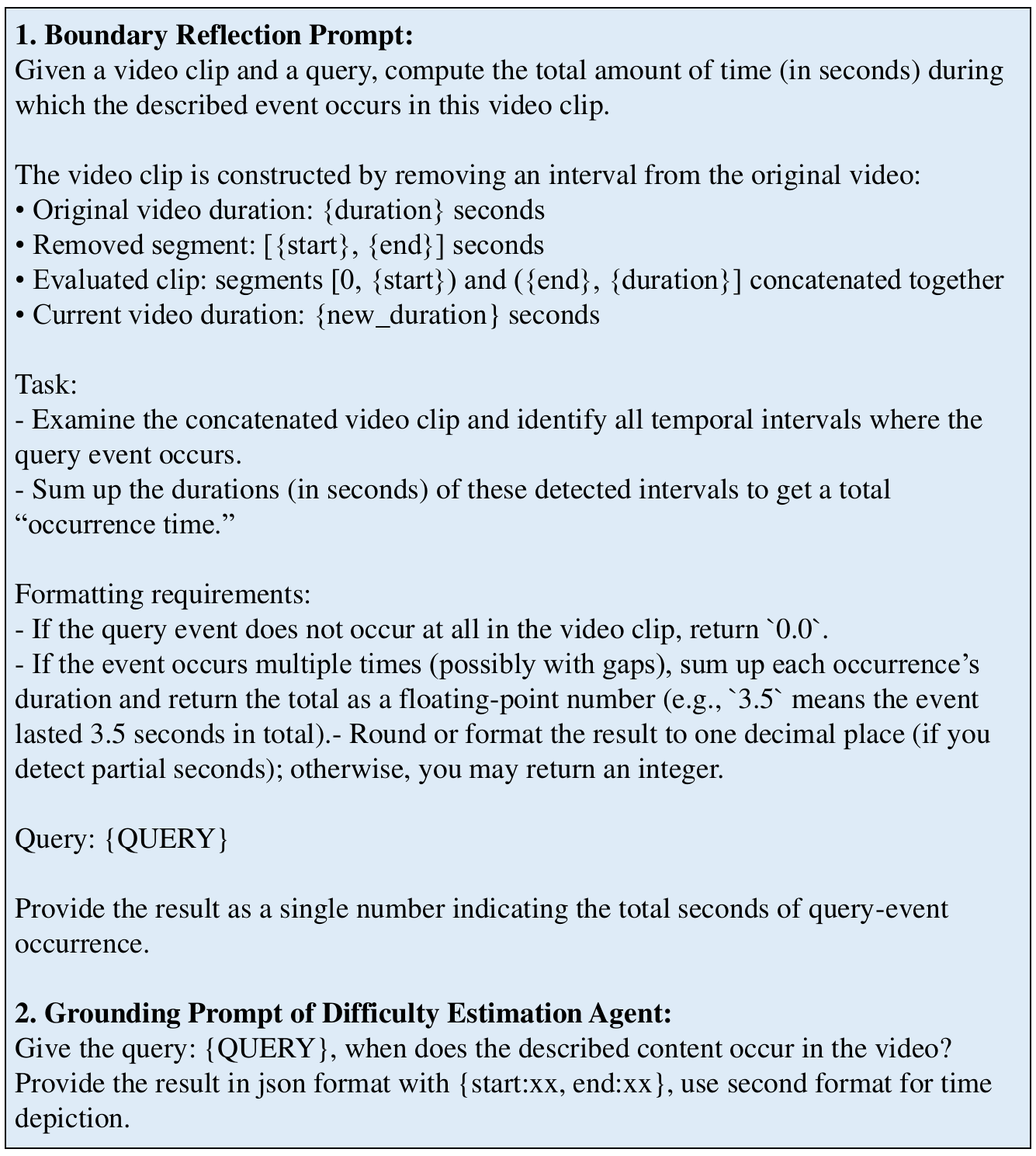} % 调整图片宽度
    \vspace{-4mm}
    \caption{Prompts used in \method.
    }
    \label{fig:prompt} % 为图片设置引用标签
\end{figure}

\subsubsection{Prompt Templates}
We present the BRA and DEA prompts utilized by \method across various operations, as shown in Fig.~\ref{fig:prompt}. Additionally, the GRPO training prompt follows the same design as VideoChat-R1~\cite{li2025videochat}. These prompts are meticulously designed to guide \method in effectively performing its subtasks, ensuring accurate responses.

\section{Conclusion}
In this work, we propose a curriculum reinforcement learning framework with reflected boundary annotations. We propose a boundary reflection agent that utilizes MLLMs to predict query-relevant timestamps outside the annotated intervals, allowing us to identify and filter out partially annotated samples.  Additionally, we employ a Difficulty Estimation Agent to assess the training difficulty of each sample and design a curriculum RL strategy that dynamically masks videos of hard-to-ground samples according to the training steps, easing the training difficulty of hard-to-ground samples and providing clearer preference during the RL training progress. Interestingly, with only 10\% of the training samples and 21\% of the computational budget, our method outperforms full-data counterparts under both GRPO and SFT paradigms.

While \method demonstrates substantial improvements in the VTG task, it still has some limitations. First, \method requires additional model inference to identify partially annotated samples and hard-to-ground samples. Secondly, our curriculum RL strategy uses a dynamic masking that considers only the temporal distance from the ground-truth segment, without accounting for the semantic relevance to the query. A potential direction is to integrate additional models to enable semantic-guided masking.

	% or
	

	%\section*{Acknowledgment}


	\ifCLASSOPTIONcaptionsoff
	\newpage
	\fi

	
	
	%\bibliographystyle{IEEEtran}
	
	% \end{thebibliography}
	{
		\small
		\bibliographystyle{unsrt}
		\bibliography{ref}
	}

	
	% \input{chapters/8-interest_of_conflict}
\end{document}